\documentclass[10pt,twocolumn,letterpaper]{article}

\usepackage{cvpr}
\usepackage{times}
\usepackage{epsfig}
\usepackage{graphicx}
\usepackage{amsmath}
\usepackage{amssymb}
\usepackage{enumitem}
\usepackage{multirow}
\usepackage{booktabs}       % professional-quality tables 
\usepackage{graphicx}
\usepackage{wrapfig}
\graphicspath{{figure/}{../figure/}}

\usepackage[pagebackref=true,breaklinks=true,letterpaper=true,colorlinks,bookmarks=false]{hyperref}

 \cvprfinalcopy % *** Uncomment this line for the final submission

\def\cvprPaperID{525} % *** Enter the CVPR Paper ID here

\ifcvprfinal\fi
\begin{document}

%%%%%%%%% TITLE
\title{Image Restoration Using Deep Regulated Convolutional Networks}

\author{Peng Liu\\
University of Florida, USA\\
{\tt\small pliu1@ufl.edu}
\and
Xiaoxiao Zhou \\
Sun Yat-Sen University, China\\
{\tt\small zhouxx5@mail2.sysu.edu.cn}
\and
Yangjunyi Li\\
University of Florida, USA\\
{\tt\small yangjunyili92@ufl.edu}
\and
El Basha Mohammad D \\
University of Florida, USA\\
{\tt\small mdelbasha@ufl.edu}
\and
Ruogu Fang \\
University of Florida, USA\\
{\tt\small ruogu.fang@bme.ufl.edu}
}

\maketitle
%\thispagestyle{empty}

%%%%%%%%% ABSTRACT
\begin{abstract}
While the depth of convolutional neural networks has attracted substantial attention in the deep learning research, the width of these networks has recently received greater interest~\cite{Zagoruyko2016WideRN, lu2017expressive, DBLP:journals/corr/LiuF17}. The width of networks, defined as the size of the receptive fields and the density of the channels, has demonstrated crucial importance in low-level vision tasks such as image denoising and restoration~\cite{DBLP:journals/corr/LiuF17}. However, the limited generalization ability, due to increased width of networks, creates a bottleneck in designing wider networks. In this paper we propose Deep Regulated Convolutional Network (RC-Net), a deep network composed of regulated sub-network blocks cascaded by skip-connections, to overcome this bottleneck. Specifically, the Regulated Convolution block (RC-block), featured by a combination of large and small convolution filters, balances the effectiveness of prominent feature extraction and the  generalization ability of the network. RC-Nets have several compelling advantages: they embrace diversified features through large-small filter combinations, alleviate the hazy boundary and blurred details in image denoising and super-resolution problems, and stabilize the learning process.  Our proposed RC-Nets outperform state-of-the-art approaches with large performance gains in various image restoration tasks while demonstrating promising generalization ability. \textit{The code is available at \url{https://github.com/cswin/RC-Nets}}.
\end{abstract}
\begin{figure}[t]
\label{fig:rc-block}
  \centering
  \includegraphics[width=0.45\textwidth]{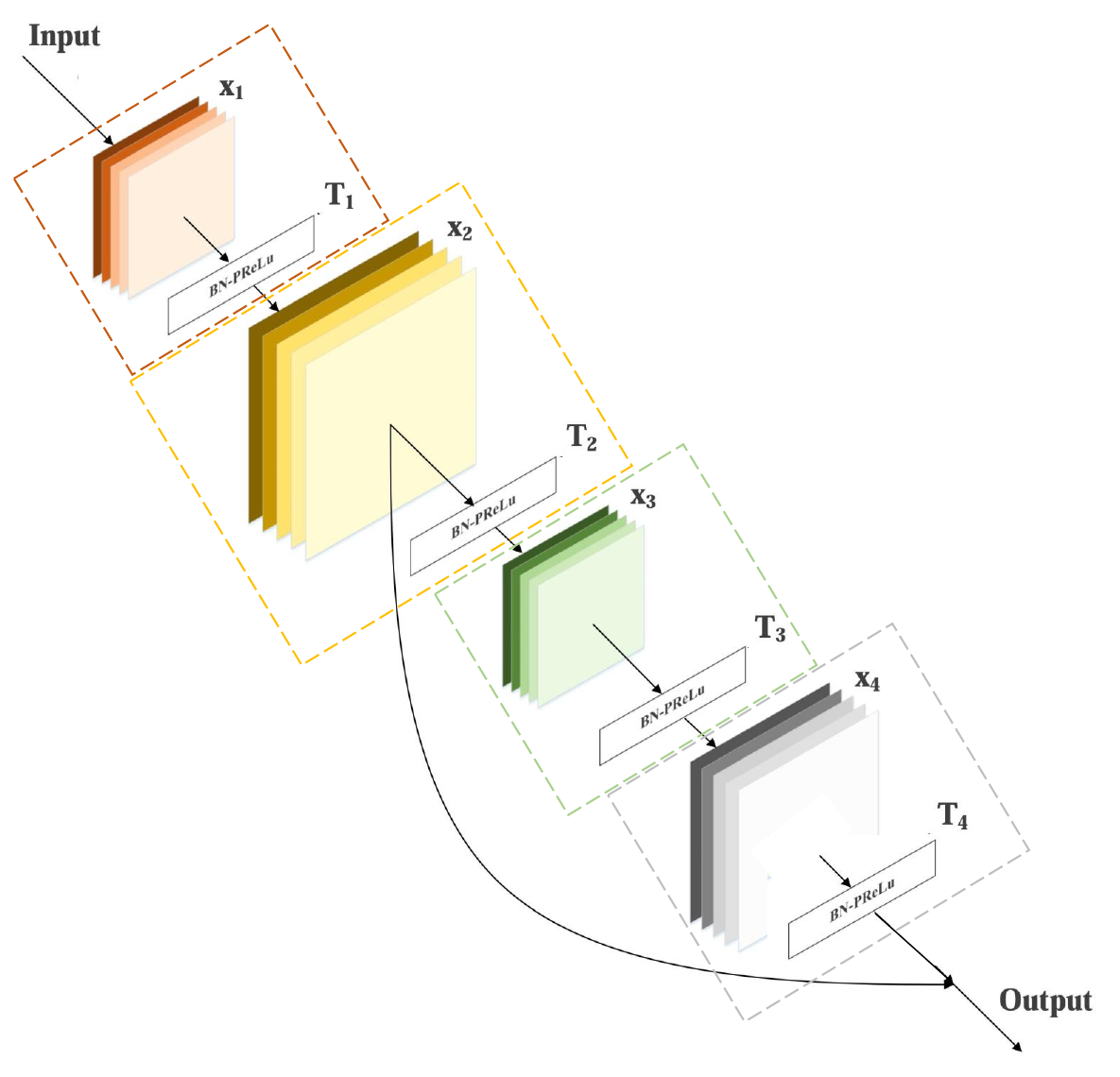}
  \caption{A regulated convolution block with 4 composite units (dotted boxes). The first and third group-squares present $1\times1$ convolution; the second and last ones indicate large and small convolution, respectively. The large convolution is regulated by the small one. }
% \label{fig:rc-block}
\end{figure}
%%%%%%%%% BODY TEXT
\section{Introduction} \label{intro}
Image restoration aims at recovering a high-quality image from a corrupted one. In particular, image denoising and single image super-resolution (SR) are two of the most important tasks. Recent works in both image denoising and SR consist mainly of learning-based methods~\cite{dong2014learning,kim2016accurate,zhang2016beyond,mao2016image,DBLP:journals/corr/LiuF17}, which learn mapping between spaces of the corrupted images and the high-quality images. Among them, Wider Inference Networks (WIN)~\cite{DBLP:journals/corr/LiuF17} have demonstrated substantial performance gain in additive white Gaussian noise denoising. By adopting large reception fields (filters) and dense channels in convolutional neural networks (CNNs), WIN can accurately exploit prominent image features to infer high-quality images.

However, wider networks such as WIN suffer from low generalization ability. For example, WIN only achieves superior performance in additive Gaussian noise removal with mediocre performance on image SR. To overcome this critical constraint, we investigated the network architecture and found three inherent limitations that constrain its generalization capability.

First, large convolution filters introduce bias. While large convolution filters (e.g. $128\times7\times7$) improve the performance of extracting prominent local features (corners and edge/color conjunctions)~\cite{zeiler2014visualizing}, they simultaneously introduce bias in the network to learn specific features of similar pixel distributions ~\cite{DBLP:journals/corr/LiuF17}. The introduced ``bias'' can boost performance significantly in one single task such as Gaussian denoising, but will degrade the performance in other tasks, such as image SR, deblurring and inpainting. This is one of the reasons that most learning-based image restoration methods~\cite{kim2016accurate,zhang2016beyond,mao2016image} use smaller convolution filters (e.g. $3\times 3$).

\begin{figure}[t]
\label{fig:rcnets}
  \centering
  \includegraphics[width=0.40\textwidth]{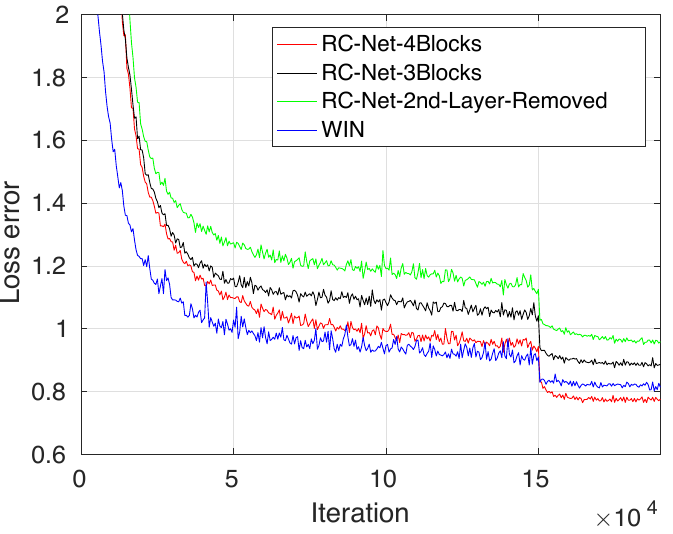}
  \caption{Comparison of validation error of RC-Nets with different number of blocks, a RC-Net having the $2^{nd}$ layer removed, and WIN during training.}
  \vspace{-1 em}
\end{figure}

Second, large convolution filters lead to significant performance fluctuations in learning (see Fig.~\ref{fig:rcnets}). The larger the convolution filters, the greater the feature variance will be, especially when noisy level is high in the training images. Small convolution filters (e.g. $3\times3$) tend to capture subtle features favoring shared weights ~\cite{szegedy2015going}. This instability issue limits the filter size in learning-based image restoration methods~\cite{mao2016image}. 

Third, large convolution filters require expensive computation. As convolution operations dominate the computation times in CNNs~\cite{he2015convolutional}, small filter sizes allow networks go deeper without significantly increasing the computational cost. Most learning-based image restoration methods~\cite{kim2016accurate,zhang2016beyond} are designed to be very deep (20 to 30 layers) by embedding $3\times3$ filters only, as the generalization capability relies on the network's depth. 

From the three limitations above, our focus is concentrating on how to balance between the generalizability and performance. 

\begin{figure*}[t]
  \centering
  \includegraphics[width=\textwidth]{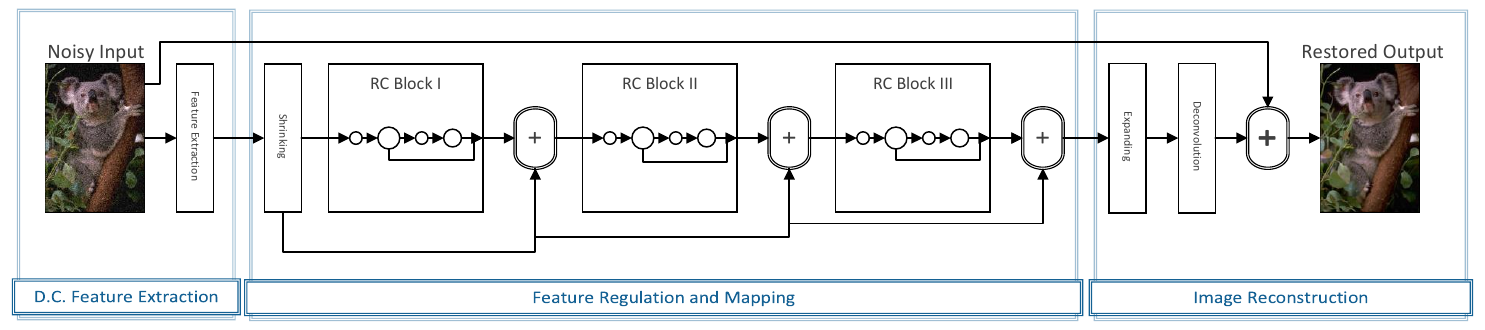}
  \caption{A deep RC-Net with four RC-blocks. In RC-blocks, the largest, medium  and smallest size of circles denote a composite unit using large, small, and $1\times1$ convolution filters, respectively. The $\bigoplus$ represents a summation computing and indicates residual learning.}
\label{fig:rc-struc}
% \vspace{-1 em}
\end{figure*}

In this work, we investigate a scalable
network structure for image restoration with both generalizability and high performance.  Inspired by WIN~\cite{DBLP:journals/corr/LiuF17}, we propose a novel cascaded scalable architecture, called Deep Regulated Convolution Network (RC-Net) (Fig.~\ref{fig:rc-struc}), which comprises of regulated sub-network blocks (Fig.~\ref{fig:rc-block}) cascaded by shortcut connections~\cite{he2016deep}, to handle both image denoising and SR.

Our proposed RC-Net addresses the aforementioned three limitations. The key contribution is to take advantages  of fusing both large and small filters, by regulating the large convolution filters by small filters. This combination of varying filter sizes can both extract the prominent image features and improve the network's generalizability at reasonable computational cost.

%  we interpolate $1\times1$ convolutional layer between other size of convolutional layers, such as the layers adopting $7\times7$ and $3\times3$ filters, except for the top two convolutional layers, which are remained for extacting the prominent features. Furthermore, we follow~\cite{dong2016accelerating} to add a transition layer with the $1\times1$ size filters at the beginning and the end of the network separately. On the one hand, it restricts feature maps to be in a low-dimensional space, and on the other hand, it can buffer the convolutional info flow between different convolutional layers to avoid the fluctuation of weight updating. 

We present comprehensive experiments on both image denoising and SR to evaluate our networks. We show that 1) Our deep regulated convolutional networks can increase
accuracy via regulated convolution, producing results
substantially better than previous networks (DnCNN~\cite{zhang2016beyond}, RED-Net~\cite{mao2016image}, and WIN~\cite{DBLP:journals/corr/LiuF17}) on image denoising; 2) Our deep regulated convolutional networks can be generalized to deal with single image SR problems, producing impressive results of high-resolution images that are more appealing for human vision. 
% The last restriction lies on embedding Bach normalization (BN)~\cite{ioffe2015batch} in each layer of WIN-net.  leads to significant performance improvement. However, BN 

%-------------------------------------------------------------------------
\section{Related Work}

\paragraph{Deep Learning for Image Restoration.}Recently, deep learning based methods have shown impressive performance on both high-level~\cite{krizhevsky2012imagenet,szegedy2015going,he2016deep} and low-level~\cite{zhang2016beyond,kim2016accurate,mao2016image,dong2014learning} vision research fields, compared to the non-CNN based models~\cite{dabov2009bm3d, gu2014weighted,timofte2014a+,huang2015single}.  ``Deeper is better'' has been considered as a design criteria of building convolutional neural networks (CNN). The preference of deep and thin CNNs stem comes from the success of deep networks in high-level vision \cite{krizhevsky2012imagenet}. Deep learning based approaches boost the performance with the cost of increasing complexity and network depth.

Despite of having high capability of nonlinear representation, deep neural networks usually encounter potential problems including gradient vanishing, over-fitting, degradation, and high inference cost. Diverse techniques have been utilized to cope with these obstacles; RED-NET~\cite{mao2016image} employs a number of skip-connections (shortcuts) to link convolution layers with mirrored deconvolution layers; VDSR~\cite{kim2016accurate} adopts very small ($3\times3$) convolution filters to restrict the amount of parameters and exploits a skip-connection to carry the input to the
end layer and reconstruct residuals, both of which contribute to speeding up training and reducing the degradation issue; DnCNN accelerates training by assembling batch normalization layers (BN)~\cite{ioffe2015batch} and $3\times3$ filters in every convolution layer, produces the residual between high-quality and corrupted images. 

\paragraph{Effective Feature Extraction and Learning.}WIN~\cite{DBLP:journals/corr/LiuF17} is built with a competitive shallow network. which is naturally superior in terms of training efficiency. WIN is stacked with 4 composite layers of three consecutive operations: convolution (Conv) layer with size $128\times7\times7$, batch normalization (BN) Layer~\cite{ioffe2015batch}, and rectified linear unit (ReLU) Layer~\cite{nair2010rectified}. In addition, a skip-connection linking the input with the end layer allows the network to predict a residual image.

With the following three key contributions, WIN offers a performance breakthrough in the Gaussian image denoising task: 1) Utilizing highly dense convolution layers composed of large filters to extract prominent local features;  2) Embedding BN layers to learn the mean and variance of feature maps associated with the pixel-distribution of training images. BN is especially helpful when pixel distributions are similar due to the same type of corruption. BN also performs as a memory storage to keep the shareable mean and variance; 3) Linking input to the end layer with a skip-connection can lead to residual learning and carrying shareable information from corrupted images to the end layers, both of which contribute to improving the accuracy of the loss-error calculation via comparing with the ground-truth images.

However, WIN also suffers from a number of limitations despite of the substantial performance margin. WIN is not able to be extended to other sub-problems of image restoration, such as SR. This is because the difference between low-resolution (LR) images and high-resolution (HR) images is subtle and not merely Gaussian distribution. Furthermore, there are fewer similarities shared by the pixel-level distribution features among the LR images. In this case, in addition to the disadvantages mentioned in Section~\ref{intro} on large filters, BN layers cannot help but restrain the network's generalizability due to data normalization. This is especially important to SR because the mapping between a subtle residual from the normalized data is no longer easier than mapping from the original data.

Convolution is the primary operation in CNNs. The number and size of filters determine the type of features extracted and the computational cost. Herein, design of the convolution filter structures is a key component of developing deep CNN models to optimize the performance. However, most deep models primarily adopt $3\times3$ convolution filters and interpolate using $1\times1$ convolution filters to reduce the number of feature-maps, such as bottleneck blocks used in ResNet~\cite{he2016deep} and DenseNet~\cite{huang2016densely}. To challenge the current trend, we propose to regulate the large filters with smaller ones to achieve a balance between performance generalizability, which remarkably outperform the state-of-the-art using only small filters and bottleneck connections.

\section{Regulating Convolution Nets} \label{sec:method}
Consider a corrupted image $x_{0}$ is passed through a convolutional network. The network intends to learn a mapping function $F$ between the corrupted image $x_{0}$ and a noise-free image $y$. The convolutional network contains $L$ convolution layers (Conv), each of which implements a feature extraction transition:
\begin{equation} \label{eq:1}
y_{l}=Conv(x_{l}, f_{l},n_{l},c_{l})  
\end{equation}
where $l$ indexes the layer, $x_{l}$ denotes the $l$'s input, and $f_{l}$, $n_{l}$, and $c_{l}$ represent the filter size, filter number, and channel number, respectively. $y_{l}$ are the feature-maps extracted from $x_{l}$ by $Conv(\cdot )$.  

As the top and bottom layers have different functional attentions~\cite{zeiler2014visualizing}, the network can be decomposed into three parts (see Fig.~\ref{fig:rc-struc}): densely convolved feature extraction, feature regulation and mapping, and image reconstruction.  

\paragraph{Densely convolved feature extraction.}We use a considerable amount of large filters in the first two~\cite{zeiler2014visualizing} convolutional layers \textit{densely convolutional feature-extraction layers} to extract diverse and representative features for feature mapping and spatial transformation. This part is similar to WIN, but WIN adopts dense convolutions in all layers and does not distinguish between the hierarchical characteristics of the bottom-up layers. Based on the empirical results of WIN, we define densely convolved features extracted from the $l^{th}$ layer as:
\begin{equation} \label{eq:2}
y_{l}=Conv(x_{l}, f_{l},n_{l},c_{l})_{f\geqslant 7\times7,n\geqslant 128}
\end{equation}
% where $d$ denotes a densely convolutional transformation. 

\paragraph{Composite unit.}Motivated by~\cite{he2016deep,zhang2016beyond,dong2016accelerating}, we combine a convolution (Conv) layer, a batch normalization (BN) layer~\cite{ioffe2015batch}, and a Parametric Rectified Linear Unit (PReLU) layer~\cite{He_2015_ICCV} as a composite unit in our proposed RC-Net, except for the last layer, which is a single deconvolutional layer. Following~\cite{huang2016densely}, we define the mapping of the composite unit as a composite function $T \left ( \cdot  \right )$:
\begin{equation} \label{eq:3}
T \left ( \cdot  \right )=PReLU(BN(Conv(\cdot ))) 
\end{equation}
\begin{equation} \label{eq:4}
T_{l} \left (x_{l}  \right )=PReLU(BN(y_{l-1})) 
\end{equation}
% where $PReLU$ represents a non-linear mapping function, and $BN$ is feature normalization by BN. 

\paragraph{Feature regulation and mapping.} We divide the network into multiple regulated convolution \textit{RC-blocks}, which are cascaded to perform feature extraction, mapping, and transformation. We assume a RC-Net contains $m$ RC-blocks, each of which comprises of four composite units; large and small convolution filters and two other $1\times1$ convolution filters for reducing the feature map dimension and regulating features extracted from the previous composite or RC-blocks. Note, the $1\times1$ convolution filters cannot change but only exert combing effect to the features generated from the preceding composite.

The key to a RC-block is to use smaller convolution filters to regulate the feature outputs from the proceeding larger filters. This regulation processing aims to balance the feature extraction so that the features generated from the larger filters can fuse information from both the larger receptive fields and the smaller ones for finer details. 

To ensure a RC-block having sufficient representative capability for feature mapping, we also suggest the  composite unit embedding larger filters in each RC-block to adopt $n\geqslant \frac{ n^{d} }{2}$ filters with size of $f\geqslant f^{d}\times f^{d}$, where $n^{d}$ and $f^{d}$ denote the filter number and the filter size in densely convolutional feature-extraction layers, respectively. 

\paragraph{Residual learning.} Without residual learning, traditional convolutional networks pass the output of the $l^{th}$ layer as the input to the $\left ( l+1 \right )^{th}$ layer ~\cite{krizhevsky2012imagenet}. If $l$ represents a composite unit, there is a transition: $y_{l}=T_{l}\left ( y_{l-1} \right )$. Residual learning~\cite{he2016deep} can adopt a skip-connection (shortcut) to perform an identity mapping: 
\begin{equation} \label{eq:5}
y_{l}=T_{l}\left ( y_{l-1} \right )+ y_{l-1}
\end{equation}
RC-Net adopts residual learning in three different locations. Within each RC-block, there is a skip-connection linking the output of larger filters to the end of the block to fuse the information generated from the smaller filters and forms residual learning as: 
\begin{equation} \label{eq:6}
y_{l_{f_s}}=T_{l_{f_s}}\left ( y_{l_{f_l}} \right )+ y_{l_{f_l}}
\end{equation}
where $l_{f_s}$ and $l_{f_l}$ denote the layer using small and large filters, respectively. The second residual learning is performed by another skip-connection, which connects the outputs of two adjacent RC-blocks or connects the first RC-block and the composite unit that precedes it. Motivated by~\cite{kim2016accurate}, the third residual learning is achieved by adding the corrupted image to the output of the last layer, which is a deconvolutional layer that converts the preceding feature-maps to an image. 

\begin{table*}[h]
\label{fig:rc-layer}
  \centering
  \caption{\small Comparison of structure detail of RC-Net;  RC-3-Blocks which removes one RC-Block from RC-Net;  RC-2nd-Layer-Removed which removes  Composite unit(2) from RC-Net;  WIN consists of three parts as shown in table. number of parameters calculate the total parameter number of network.Comparison is based on image denoising  performance, average PSNR (dB) / SSIM / Run Time (s), are evaluated on the BSD200-test with noise level $\sigma = 50$.}
  \includegraphics[width=0.9\textwidth]{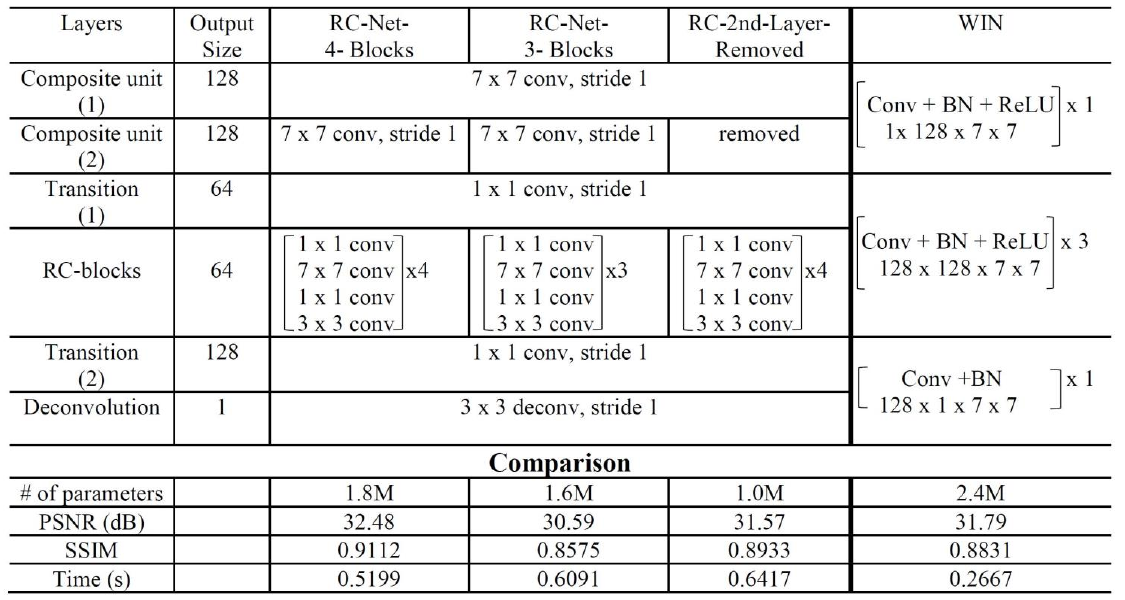}
  
\end{table*}

\paragraph{Scale controlling.}
To make RC-Nets more compact, motivated by~\cite{dong2016accelerating}, we introduce two $1\times1$ composite units as transition layers, refereed as ``Shrinking'' and ``Expanding'', which are shown on Figure~\ref{fig:rc-block}. After densely convolutional feature-extraction layers, we reduce the number of feature-maps by ``Shrinking'', in which, inspired by~\cite{lu2017expressive}, we suggest to adopt $n\geqslant \frac{ n^{d} }{2}$ filters to remain sufficient features for following RC-blocks, where $n^{d}$ denotes the filter number in densely convolutional feature-extraction layers. 

After feature regulation and mapping, we use another transition layers to expand feature-maps so that there are sufficient various features that can be provided for image reconstruction.

\paragraph{Image reconstruction.}
RC-Nets reconstruct a noise-free image $y$ by adding the input (corrupted) image $x_{0}$ to the one generated from the RC-Net's last layer: a deconvolutional layer, which can be denoted as: $Deconv(\cdot)$. Thus we can define the image reconstruction as: 
 
\begin{equation} \label{eq:de}
y=x_{0}+Deconv(x, f,n,c)_{s,n= 1}
\end{equation}
where $s, n= 1$ represents one small filters used to inverse the preceding feature-maps finely.

 %------

%------------------------------------------------------------------------

\section{Experiments and Results}
We perform experiments on both image denoising and super-resolution.

\subsection{Datasets}

\paragraph{Image Denoising.} The training data for image denoising consists of 91 images from Yang et. al.~\cite{yang2010image}.and an additional 200 images from the Berkeley Segmentation Dataset BSD200. We add Gaussian noise at four different levels ($\sigma$ = 10, 30, 50, 70) to train the RC model separately for each noise level. Noise is added using the \textit{randn} function in MATLAB. We train other comparing methods using the same dataset of 291 images.  To keep consistent with the original paper~\cite{zhang2016beyond}, DnCNN follows~\cite{chen2017trainable} to
use 400 images of size $180 \times  180$ for training.

To evaluate the denoising performance of each method, we use the BSD200 test dataset and Set12. Due to various versions of Set12, we resized images of the 12 standard image to be $481 \times 321$, as the same size as the majority of images in the BSD200 test dataset. Both datasets are used in the image denoising experiment in Table~\ref{tab:BSD-test} and Table~\ref{tab:Set12-test}. 

\paragraph{Single Image Super-Resolution.} We use the same dataset of 291 images at three different scale factors ($\times2$, $\times3$, $\times4$ ) to train the RC-Net model separately. WIN and VDSR also share the same training dataset ``291''. SRCNN~\cite{dong2014learning} uses a large ImageNet dataset~\cite{deng2009imagenet} as in its original paper~\cite{dong2014learning}. 
For evaluation, two datasets are used: BSD200~\cite{MartinFTM01} from the Berkeley Segmentation Dataset~\cite{MartinFTM01} and Set14~\cite{zeyde2010single}.

\subsection{Training}

%-------------------------------------------------------------------------

% \paragraph{Experimental Environment.}
Our RC-Net has four RC-blocks for all datasets. RC-Net has a depth of 21 layers, where 20 of them are composite units (see Section~\ref{sec:method}). The remaining layer is a deconvolutional layer (Deconv). To avoid the ``dead features''  ~\cite{zeiler2014visualizing} issue in ReLU~\cite{he2015delving}, we adopt the Parametric Rectified Linear Unit (PReLU)~\cite{he2015delving} for the activation function after each bach normalization layer (BN)~\cite{ioffe2015batch}. 

All models adopt Stochastic Gradient Descent (SGD) with mini-batch size of 64 for image denoising and 128 for SR. For model optimization, we introduced the BN layer to reduce the internal covariate shift leading to an accelerated learning speed. For weight initialization, we follow the method described in~\cite{he2015delving}. This is especially suitable for the networks adopting ReLU or PReLU.

%---

%----
%----
All experiments are trained over 50 epochs (5,000 iterations per epoch) totaling 250,000 iterations. Learning rate is initially set to 0.1 and is divided by 10 after every $15\times 10^{4}$ iterations. Weight decay is set to 0.0001 and momentum to 0.9. We follow the practice in~\cite{zhang2016beyond,mao2016image,kim2016accurate} for both image denoising and SR. Data-augmentation is performed to increase the size and variance of training samples. A $41\times 41$ crop with stride 14 is randomly sampled from an image or a horizontally flipped copy to generate image patches for additional training samples. All training is processed on 8 GeForce GTX TITAN Xp GPUs.

%-------
\begin{table*}[!]
%  \scriptsize
 \small
 \caption{ The average results of PSNR (dB) / SSIM / Run Time (seconds) of different methods on the BSD200-test~\cite{MartinFTM01} (200 images).All methods are applied on several noise level($\sigma$ = 10, 30, 50, 70).The best results are highlighted in bold.}
\label{tab:BSD-test}
\centering

  \begin{tabular}{llllll}
  \toprule
  \multicolumn{6}{c}{PSNR (dB) / SSIM / Time (s) }                   \\
    \cmidrule{1-6}
    $\sigma$ & BM3D \cite{dabov2009bm3d}   & RED-Net \cite{mao2016image} & DnCNN \cite{zhang2016beyond} &  WIN~\cite{DBLP:journals/corr/LiuF17} &  RC-Net\\
     
    \midrule
    10  & 34.02/ 0.9182/ 1.01 & 32.96/ 0.8963/ 60.73 & 34.60/ 0.9283/ 13.49  & 35.83/ 0.9494/ 28.72  & \textbf {36.36/ 0.9541}/ 32.55     \\
    30  & 28.57/ 0.7823/ 1.23 & 29.05/ 0.8049/ 60.27 & 29.13/ 0.8060/ 13.48  & \textbf {33.62}/ 0.9193/ 31.94  &  33.57/ \textbf {0.9271}/ 33.52   \\
    50  & 26.44/ 0.7028/ 2.02 & 26.88/ 0.7230/ 66.62 & 26.99/ 0.7289/ 12.55  & 31.79/ 0.8831/ 23.77  & \textbf{32.48/ 0.9112}/ 33.52      \\
    70  & 25.23/ 0.6522/ 1.98 & 26.66/ 0.7108/ 66.99 & 25.65/ 0.6709/ 13.42  & 30.34/ 0.8362/ 23.90  & \textbf{31.17/ 0.8795}/ 32.81      \\
    \bottomrule
    
  \end{tabular} 
  \vspace{0.15 em}
\end{table*}
\medskip

\begin{table*}[!]
% \scriptsize
  \small
  \caption{ The average PSNR(dB) / SSIM / Run Time (seconds)  of different methods on the resized 12 standard testset (12 images). All methods are applied on several noise level($\sigma$ = 10, 30, 50, 70). The best results are highlighted in bold.}
  \label{tab:Set12-test}
  \centering
  \begin{tabular}{llllll}
    \toprule
  \multicolumn{6}{c}{PSNR (dB) / SSIM / Time (s)}                   \\
    \cmidrule{1-6}
    $\sigma$ & BM3D \cite{dabov2009bm3d}   & RED-Net \cite{mao2016image} & DnCNN \cite{zhang2016beyond} &  WIN~\cite{DBLP:journals/corr/LiuF17}  &  RC-Net\\
     
    \midrule
    10  & 34.28/ 0.9208/ 0.91  & 30.48/ 0.8610/ 61.89   & 33.30/ 0.9096/ 12.79  &  37.24/ 0.9546/ 12.03  & \textbf {38.54/ 0.9627}/ 29.56\\
    30  & 29.09/ 0.8287/ 0.99  & 30.48/ 0.8610/ 62.71   & 28.19/ 0.8151/ 12.52  &  34.89/ 0.9162/ 12.77  & \textbf {35.18/ 0.9409}/ 29.50\\
    50  & 25.23/ 0.6522/ 1.98  & 28.03/ 0.7988/ 63.66   & 25.85/ 0.7522/ 12.48  &  32.99/ 0.8825/ 12.37  & \textbf {34.55/ 0.9270}/ 30.44\\
    70  & 25.20/ 0.7149/ 1.75  & 27.95/ 0.7950/ 63.20   & 23.75/ 0.6890/ 12.41  &  30.91/ 0.8263/ 11.75  & \textbf {32.87/ 0.8961}/ 30.43\\
    \bottomrule

  \end{tabular}
  \vspace{0.15 em}
\end{table*}

\begin{figure*}[!]
 \vspace{-1em}
  \centering
  \includegraphics[width=\textwidth]{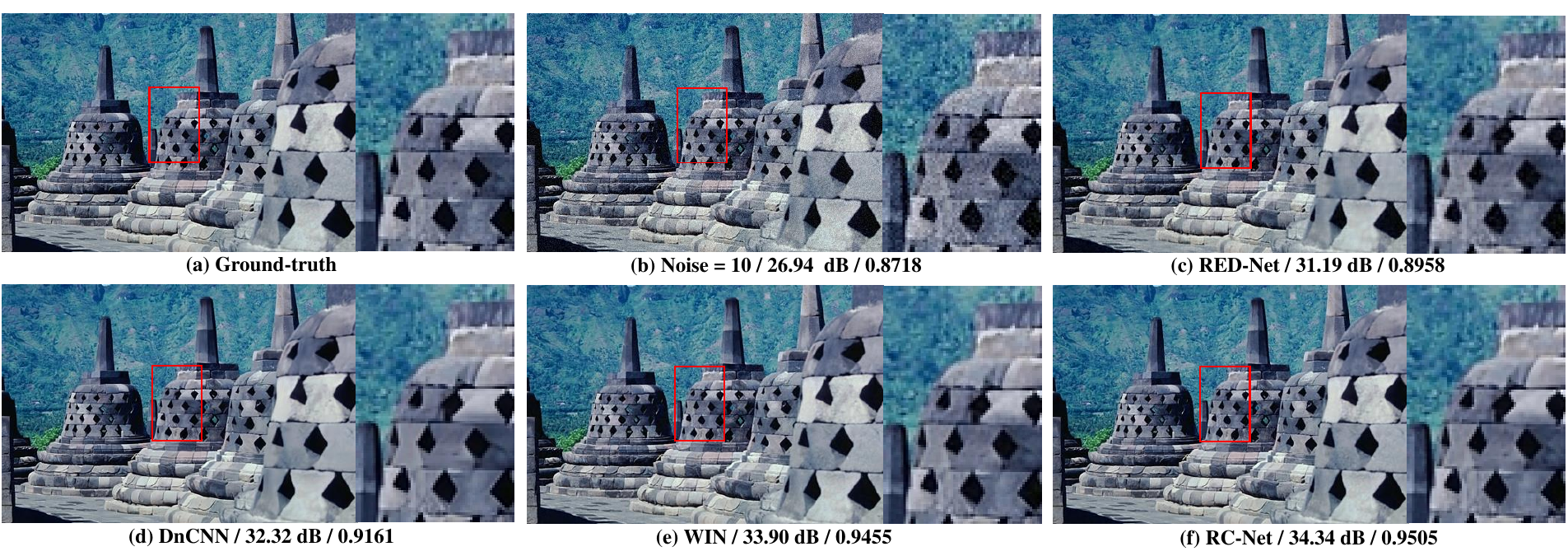}
  \caption{\small Visual results of one image from BSD200-test with $\sigma=10$
  along with PSNR(dB) / SSIM.}
  \vspace{0.3 em}
\label{fig:N10}
\end{figure*}

\begin{figure*}[!]
 \vspace{-1em}
\centering
  \includegraphics[width=\textwidth]{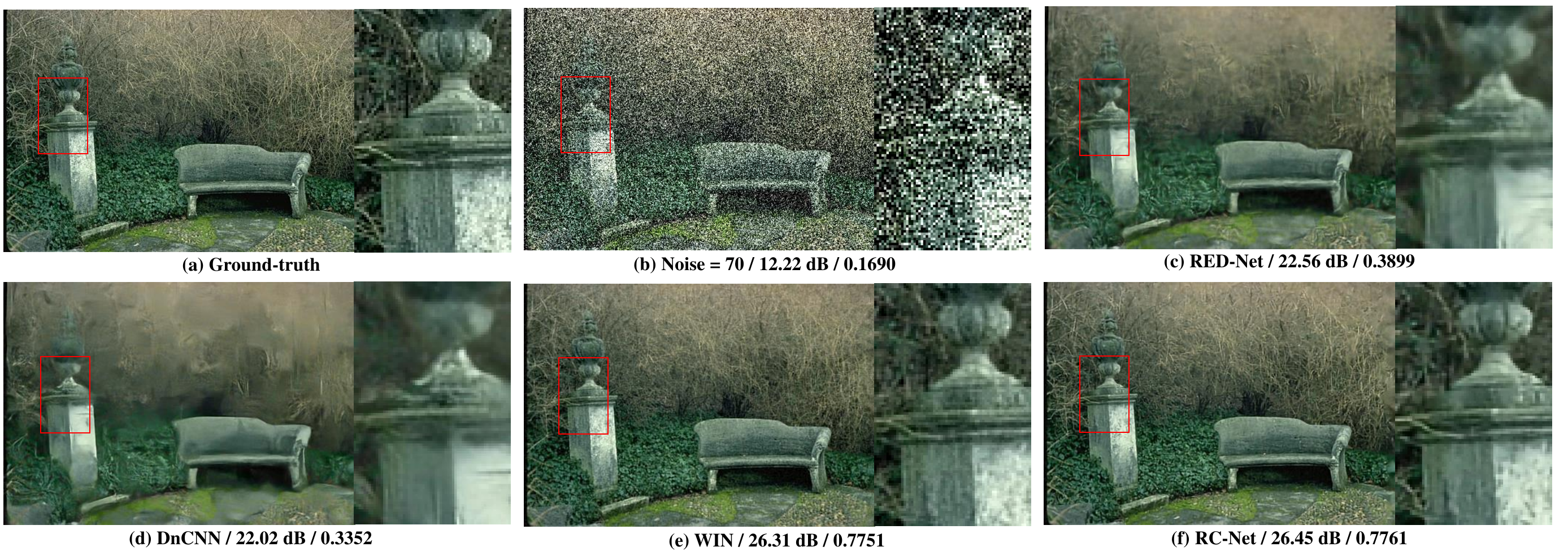}
  \caption{\small Visual results of one image from BSD200-test with $\sigma=70$
  along with PSNR(dB) / SSIM.}
  \vspace{0.1 em}
\label{fig:N70}
\end{figure*}
%-------------------------------------------------------------------------
\subsection{Image Denoising}
Both quantitative and qualitative comparisons of various denoising and SR methods with our proposed RC-Net are provided. In Table~\ref{tab:BSD-test} and Table~\ref{tab:Set12-test}, we provide quantitative evaluations of five different methods: BM3D~\cite{dabov2009bm3d}, RED-Net~\cite{mao2016image}, DnCNN~\cite{zhang2016beyond}, WIN~\cite{DBLP:journals/corr/LiuF17}, and RC-Net. PSNR, SSIM, and run time are used as evaluation criteria.

From the results, our proposed RC-Net method significantly outperforms BM3D, RED-Net, and DnCNN on both datasets. When compared with WIN, our RC-Net method has relatively similar performance when noise level is 30. For example, on the dataset BSD200, our RC-Net method generates a PSNR of 33.57 dB while WIN generates 33.62 dB. While RC-Net and WIN have extremely close PSNR, RC-Net still outperforms WIN in terms of SSIM. With the increased noise level, the denoising results of RC-Net surpasses WIN remarkably. For example, on Set12, RC-Net generates an average PSNR of 32.87 dB, while WIN produces a PSNR of 30.34 dB. 

We illustrate visual comparisons of four different methods: RED-Net, DnCNN, WIN, and RC-Net in Figure~\ref{fig:N10} and Figure~\ref{fig:N70} with sample images from BSD200 dataset. In Figure~\ref{fig:N10}, our RC-Net method excels in maintaining the  detailed particle structure on the building’s surface while other methods fail to keep this important texture information. Figure~\ref{fig:N70} is a challenging image due to a great quantity of plants with the same color and the high noise level. In this challenging case, our method still excels in restoring the pillar’s texture features, as well as clear contours and details of the chair and branches. 
%As the same as figure~\ref{N50_v3}, horn and face of rhinoceros are much more clear and sharp than previous methods. Figure~\ref{N70_v3}is a challenging image due to great quantity of plants in the same color and high-level noises. The RC recovered image shows clear contours and details of pillar, chair, and branches. 

\begin{figure*}[ht]
\centering
  \includegraphics[width=0.9\textwidth]{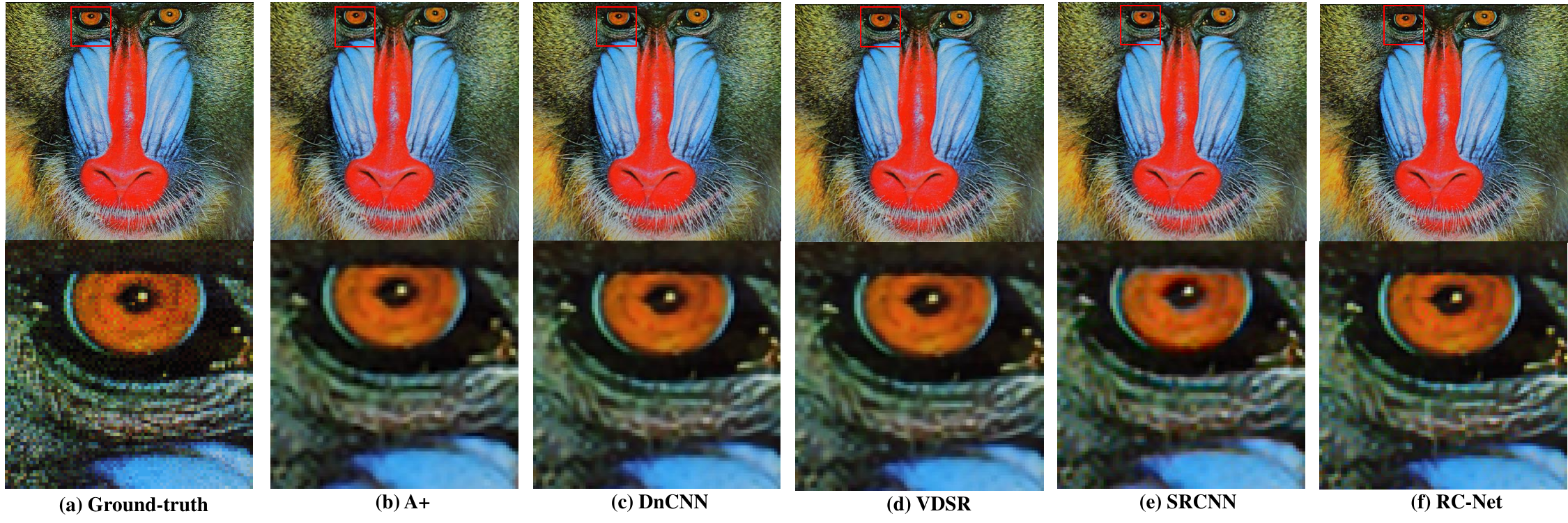}
  \caption{Visual results of one image from Set14 with scale factor x2 along with PSNR(dB) / SSIM.}
  % \vspace{-1em}
\label{fig:S2}
\end{figure*}

\begin{figure*}[!]
\centering
  \includegraphics[width=0.9\textwidth]{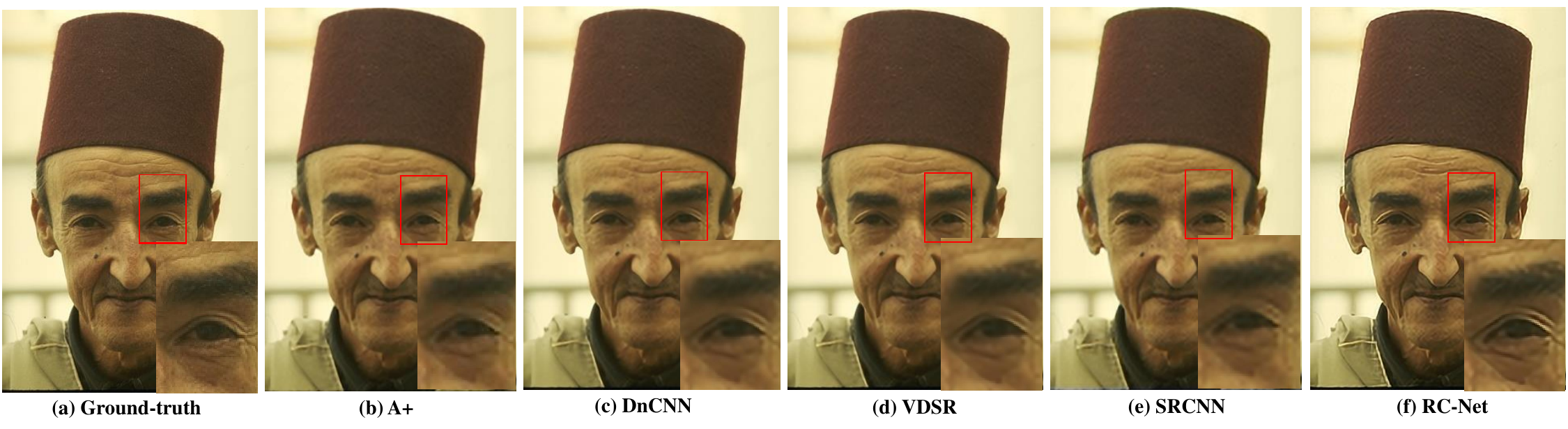}
  \caption{Visual results of one image from B100 with scale factor x3 along with PSNR(dB) / SSIM.}
\label{fig:S3}
\end{figure*}

The visual results show a well match with quantitative results. Our method has improved performance compared to the state-of-the-art image denoising approaches, especially when the noise level is high. Our RC-Net gains the best SSIM term among all test datasets as well as the best human visual image result.

\begin{figure*}[!]
\centering
  \includegraphics[width=0.9\textwidth]{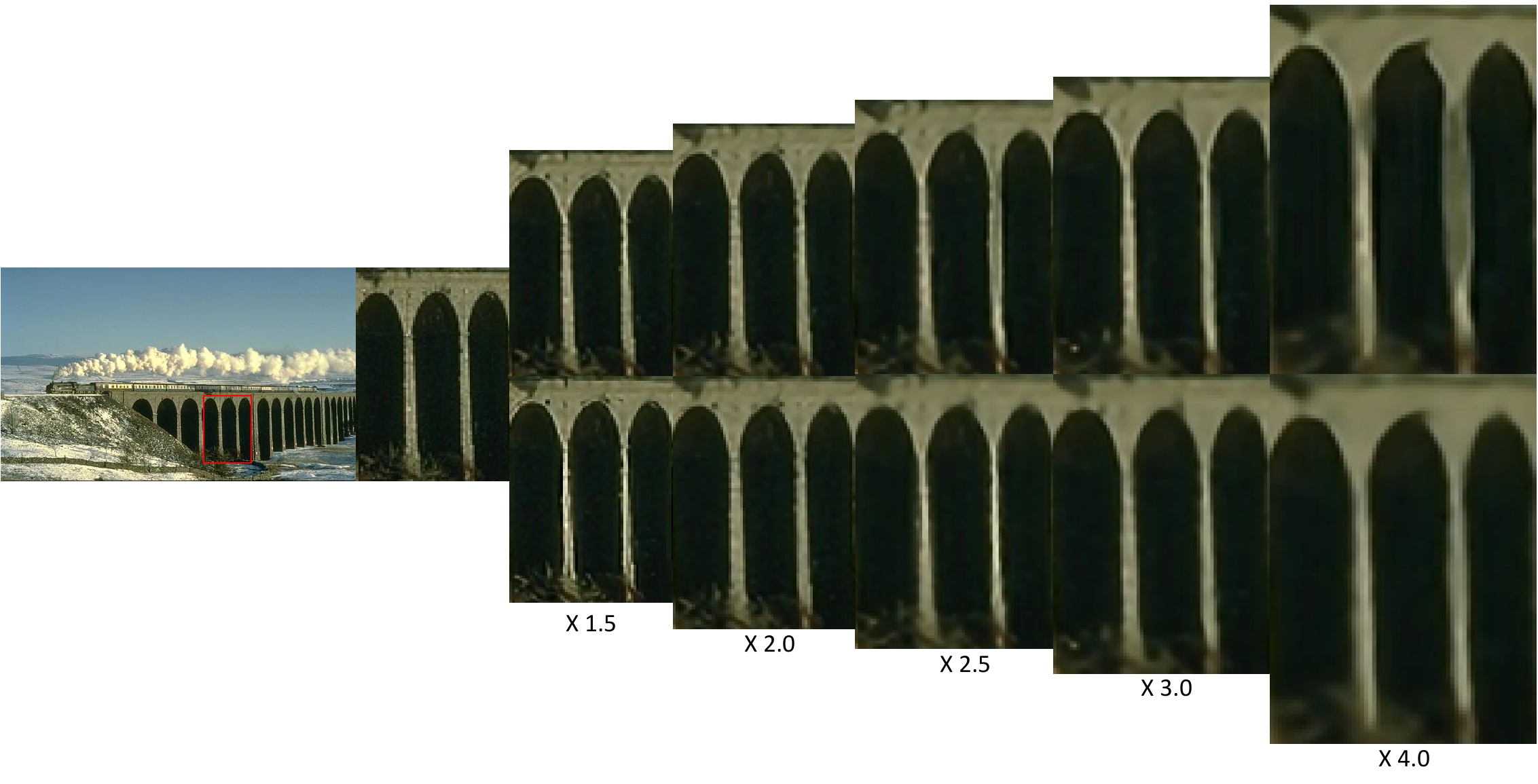}
  \caption{Visual results of single network for multiple scale factors.  ROIs are selected and enlarged on the right. \textbf{Top}: Restoration results within RC-blind network, RC-blind is trained on scale factor = $\times2$, $\times3$, and $\times4$.  \textbf{Bottom}: SR results generated by VDSR}
\label{fig:Sb}
\end{figure*}

% \begin{table}
% \begin{center}
% \begin{tabular}{|l|c|}
% \hline
% Method & Frobnability \\
% \hline\hline
% Theirs & Frumpy \\
% Yours & Frobbly \\
% Ours & Makes one's heart Frob\\
% \hline
% \end{tabular}
% \end{center}
% \caption{Results.   Ours is better.}
% \end{table}

%-------------------------------------------------------------------------
\subsection{Super-Resolution}
To demonstrate the generalizability of RC-Net, we show illustrations of single image super-resolution using our RC-Net and other state-of-the-art approaches (A+~\cite{timofte2014a+}, SRCNN~\cite{dong2014learning}, VDSR~\cite{kim2016accurate}, and DnCNN~\cite{zhang2016beyond}) in Figure~\ref{fig:S2} and Figure~\ref{fig:S3}. When we scale an image with $2\times$ factor as shown in Figure~\ref{fig:S2}, it is obvious that our method obtains superior eyes texture and boundary of baboon compared to the state-of-the-art methods. In Figure~\ref{fig:S3}, our method has demonstrated notable capability to maintain the image details in SR by capturing the wrinkle and eyebrow texture while the comparing state-of-the-art methods, which tend to lose the texture information and result in a blurred image. 

Furthermore, most existing methods use multiple networks to handle different scale factors in image SR, while one general network that can perform SR at various scale factors is preferred. Figure~\ref{fig:Sb} exhibits the ability of a single RC network (RC-blind) to perform SR at different scale factors. We compare the performance of recovered images at multiple scales using RC-blind (top) with VDSR (bottom). From Figure~\ref{fig:Sb}, we can observe that while both methods can handle different scale factors, our proposed RC-Net outperforms VDSR in terms of clarity and sharpness.

\section{Discussion}

%  \begin{table}[ht]
% \scriptsize
% \caption{\small Comparisons of the performance and scale of different RC-Nets with WIN. The ``RC-Net-3/4-Blocks'' represents a RC-Net has 3/4 RC-blocks. The ``RC-2nd-L-Removed''  means the second layer of a RC-Net is removed. The image denoising performance, average PSNR (dB) / SSIM / Run Time (s), are evaluated on the BSD200-test with noise level $\sigma = 50$.}
% \label{tab:compare RCs}
%  \centering
%   \begin{tabular}{lllll}

%     \toprule
%      Model  & PSNR(dB)   & SSIM &  Scale(Mbyte) & Time(s) \\
     
%     \midrule
%     RC-4-Blocks  & 32.48 & 0.9112 & 7.0  & 0.5199     \\
%     WIN & 31.79 & 0.8831 & 9.3  & 0.2667      \\
%     RC-2nd-L-Removed & 31.57 &  0.8933 & 3.9  & 0.6417   \\
%     RC-3-Blocks & 30.59 & 0.8575 & 6.1  & 0.6091    \\

%     \bottomrule
%   \end{tabular}
%   \vspace{-1em}

% \end{table}

Ostensibly, the specific structure design on RC-Nets (Table~\ref{fig:rc-layer}) that differs from other deep convolutional networks merely in the cascaded RC-Blocks, which include large-small filter combination for regulation (Figure~\ref{fig:rc-block}). However, this small modification leads to a remarkable image restoration performance of the networks.

\paragraph{Diversified Feature Representation.} The most direct consequence of introducing RC-Blocks is efficiency in diversified feature extraction . Table~\ref{fig:rc-layer} shows the comparison between RC-Nets and WIN, which is a simplified structure of RC-Net without RC-blocks but more parameters. It is apparent that RC-Nets outperforms WIN even with with a much smaller parameter number (1.8 M vs. 2.4 M). 
Even though the performance degrades a little after we remove one RC-block, RC-3-Blocks Nets with only 1.6 M  parameters still outperforms state-of-the-art methods and similar to WIN. We also further remove one Composite Unit (Table~\ref{fig:rc-layer} and shows that with a minor degradation of performance, the parameters number decrease from 1.8 M to 1.0 M. These variations and comparisons demonstrate the effectiveness of our RC-Block in feature utilization and the substantial potential for a more compact network.

\paragraph{Handling Training Fluctuation.}  There is connection between filter sizes and loss-error curve as shown in Figure~\ref{fig:rcnets}. WIN shows strong fluctuating even in platform period. However, RC-Nets substantially improve the stability in the loss error during the training process and outperforms WIN after around $15\times10^4$ iterations when the entire dataset has been traversed.

\paragraph{Enhanced Generalizability.} Cascaded RC-Blocks overcomes the limitation in generalizability of merely using dense convolutional layers in WIN. RC-Nets show superb visual results in single image super-resolution (Figure~\ref{fig:S2} and Figure~\ref{fig:S3}). The collaborative effect large-small filter combination in RC-Blocks encourages feature extraction with detail preservation as both dominant and detailed features are collected by large and small filters. The overall performance of RC-Nets in both image restoration indicates the enhanced generalization ability with cascaded and specially RC-Blocks. It also implies the great potential of our proposed RC-Nets in solving broader image restoration tasks.

%------------------------------------------------------------------------
\section{Conclusion}
We proposed a new convolutional network architecture,
referred as Deep Regulated Convolutional Network
(RC-Net), for image restoration. It introduces large-small filter combinations (RC-Block) where small convolution filters regulate the features extracted by the large ones. In our experiments, we demonstrated that RC-Nets achieved state-of-the-art performance across several highly competitive datasets for multiple image restoration tasks. Moreover, RC-Nets need substantially fewer parameters and less computational cost to outperform state-of-the-art approaches with large margins, with appealing performance in structural information preservation and generating results appealing for the human visual systems.   Owing to the regulated convolutions, RC-Nets can balance the feature extraction and generalizability by fusing information from both the representative features extracted from the larger receptive fields and the smaller textures obtained from finer details.

The advantages of RC-Net lie in the balance between large and small convolutional filters, a simple combination that naturally integrates diversified feature extraction, stability in the training process, and enhanced generalizability. This design allows regularized diversified feature representation throughout the networks, which consequently yielding more efficient and compact models with highly competitive image restoration performance. Due to their efficient and compact model for feature representation, we plan to investigate the generalization ability of RC-Nets to various computer vision tasks that build on convolutional neutral networks.

% The structure of RC-Nets is scalable and compact. Comparing with WIN, it has less model scale while still outperforms WIN. By regulating the convolutional process, RC-Nets utilize parameters more efficiently. Because of regulated internal feature representations by fusing large and small filters' convolutions, RC-Nets may be more compact by finding a better combination of large and small filters. Also, RC-Nets may be extended to various computer vision tasks based on feature extraction by convolutional features. We plan to investigate RC-Net's generalization on more vision fields. 

{\small
\bibliographystyle{ieee}
\bibliography{refs}
}

\newpage
 
\appendix
% \counterwithin{figure}{section}

\section{Shortcomings of using Batch Normalization}

%  \section*{A. Shortcomings of using Batch Normalization}
A RC-Net using batch normalization (BN) does not perform as well in Super-resolution (SR) as it does in the Gaussian denoising task. On one hand, following the feature extraction in the regulated convolution layers, BN is able to enhance the high-frequency features (lines, edges, corners, etc.)  by summarizing mean and variance of feature-maps during training into two learned parameters. The enhanced features benefit both image denoising and single image SR. On the other hand, for Gaussian denoising, because the noisy images fed into networks follow Gaussian pixel-distribution, BN normalizes the information flow and keeps the distribution of each layer's inputs the same as the original input (noisy images) during training. Nevertheless, for SR, the information processed by BN is shifted and largely different from the original input (low-resolution images). This gives rise to negative effects on SR when the predicated high-resolution image is generated from the summation of the shifted signals and the original one through a skip-connection linking the input to the end of the last layer. The high-frequency features still can be maintained but the low-frequency features are latently lost during shifting and normalization. 
% as force changing the original input to be shifted and normalized, especially
This is the reason why the visual results from RC-Nets for SR are better to the human vision system but PSNR and SSIM are less than the most competitive state-of-the-art methods.

We may overcome this drawback in three potential ways: (1) Follow the work of DnCNN by directly predicting the residual between the corrupted image and the noise-free one. In this case, we remove the input-to-end skip-connection and BN layers in the first and last layers.  (2) Simply remove all BN layers from RC-Nets; (3) Dynamically route the training and testing to a specific part of a RC-Net based on the property of an input. (1) and (2) are simper and may impact the denoising performance. (3) needs an additional learning process to allow the network knows how and where to route the information. We consider (3) as a future work and implemented (2). The results for SR (see \textbf{B}) demonstrate that BN indeed has the negative effects on SR, and that RC-Nets without BN achieve competitive PSNR and SSIM with state-of-the-art methods.   
\textit{Code and models are available at \url{https://github.com/cswin/RC-Nets}}.

\section{More Results}
\begin{figure*}[h!]

 \caption{\small \textbf{Image Denoising:} Visual results of one image from BSD200-test with $\sigma=30$ along with PSNR(dB) / SSIM. Compares to state-of-the-art method, our RC-Net well recovers facial features of kids, especially the eyes are almost identical to the ground truth while state-of-the-art methods lose contours and details because of the high noise level. } 
  \includegraphics[width=\textwidth]{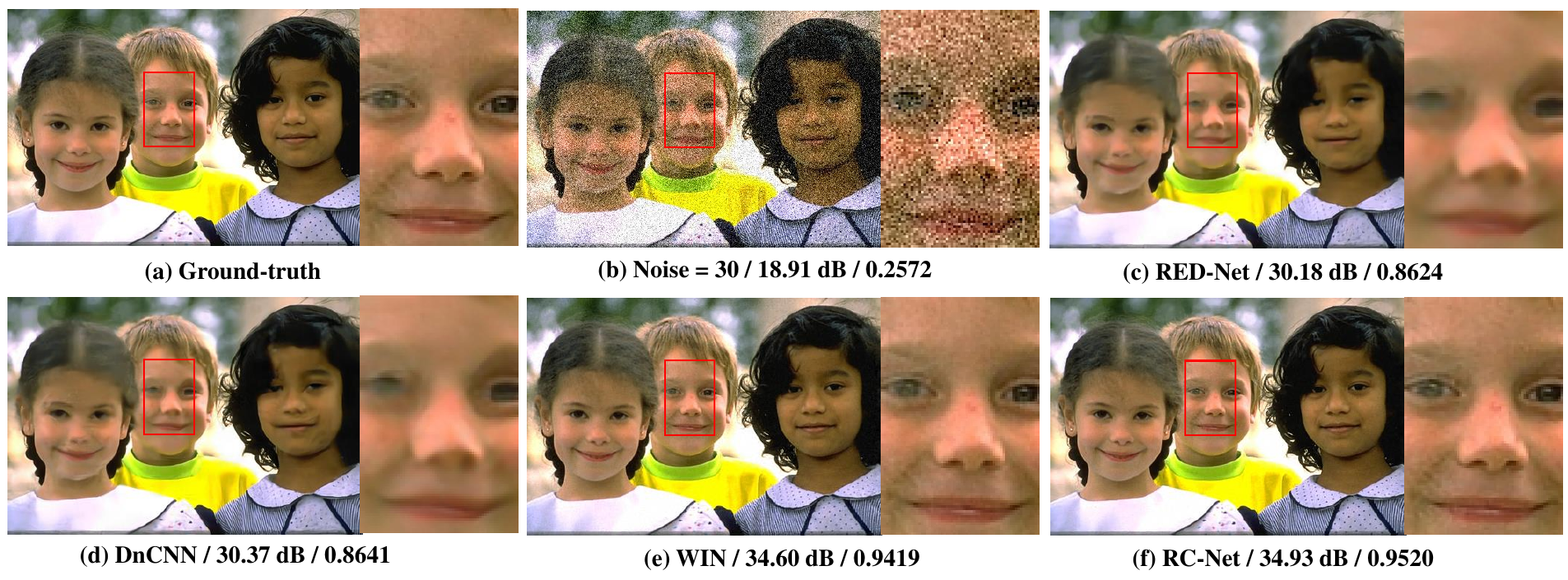}
 
\label{fig:N30}
\end{figure*}

\begin{figure*}[h!]

  \caption{\small \textbf{Image Denoising:} Visual results of one image from BSD200-test with noise level of $\sigma=50$  along with PSNR(dB) / SSIM. Horn and ears of rhinoceros are accurately reproduced using our RC-Net method. Bushes an woods in the background are discernible instead of the blurred restoration from other state-of-the-art methods. } 
  \includegraphics[width=\textwidth]{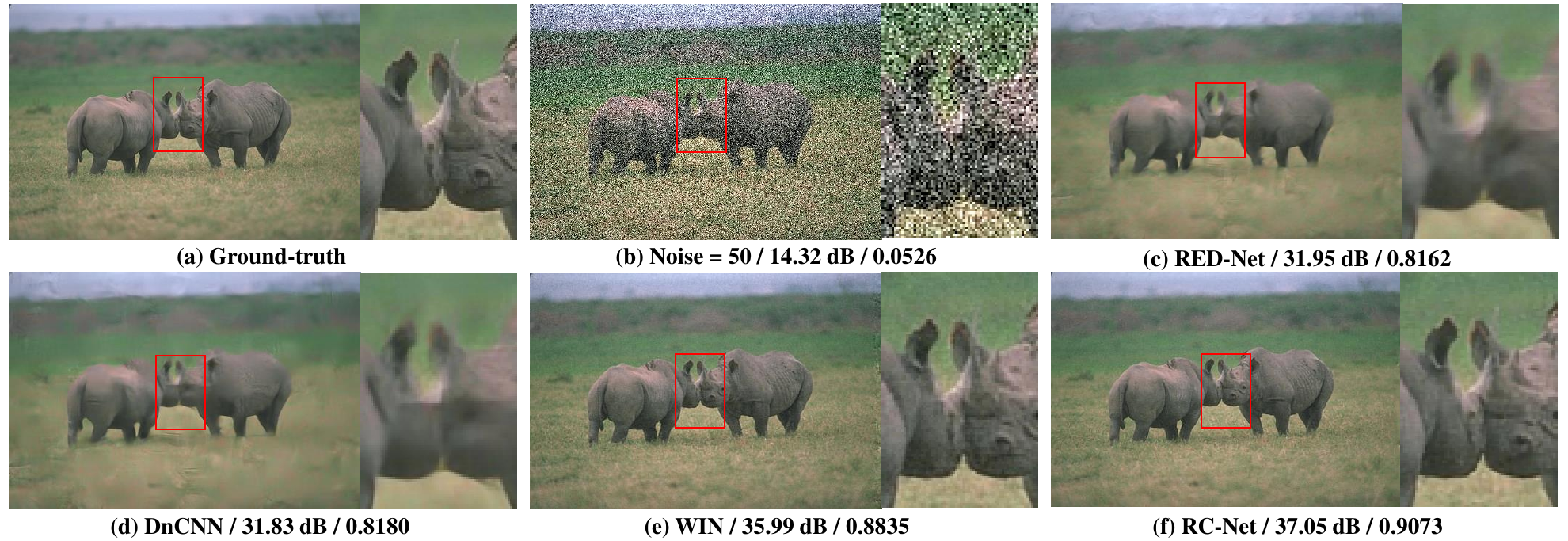}

\label{fig:N50}
\end{figure*}

More visual results are shown below to demonstrate the competitive performance of RC-Nets for image restoration. We illustrate image results on datasets BSD200-test for image denoising with noise level $\sigma=30$ (Figure~\ref{fig:N30}) and 50 (Figure~\ref{fig:N50}). While RC-Nets outperform all competing methods, they perform exceptionally well in complex images and at high noise level. Our method reconstructs well the shape and edges of left eye in Figure~\ref{fig:N30} as presented in the ROI. Similarly, in Figure~\ref{fig:N50}, contours are clean and visually discernible.

\setlength{\tabcolsep}{4pt}
\begin{table*}[h!]
  %\scriptsize
    \small
 \centering
  \caption{\small \textbf{Super-resolution:} Average PSNR / SSIM / Run Time (seconds on GPU) for scale factor $\times2$, $\times3$ and $\times4$ on datasets Set5, Set14 and B100. \textcolor{red}{Red color} indicates the best
performance and \textcolor{blue}{blue color} indicates the second best performance.}
 \label{tab:Scale}
  \begin{tabular}{llllll}
  \toprule
  \multicolumn{6}{c}{PSNR (dB) / SSIM / Time (s)-GPU }                   \\
     \cmidrule{1-6}
    Dataset & scale  & SRCNN  & VDSR  &  RC-Net with BN  &  RC-Net without BN \\
    
      \midrule
   \multirow{3}{0.8cm}{Set5} & $\times2$  & 36.55  / 0.9542 / 3.37 & \textcolor{red}{37.60} / \textcolor{red}{0.9591} / \textcolor{red}{0.13} & 35.48 / 0.9515 / 0.68  & \textcolor{blue}{37.42} / \textcolor{blue}{0.9586} / \textcolor{blue}{0.46} \\
     & $\times3$  & 32.67 / 0.9086 / 3.35 & \textcolor{red}{33.54} / \textcolor{red}{0.9207} / \textcolor{red}{0.14} & 27.47 / 0.7851 / 0.68  & \textcolor{blue}{33.43} / \textcolor{blue}{0.9191} / \textcolor{blue}{0.44}    \\
     
     & $\times4$  & 30.37 / 0.8620 / 3.30  & \textcolor{blue}{30.99} / \textcolor{red}{0.8800} / \textcolor{red}{0.15} & 26.06 / 0.7117 / 0.73  & \textcolor{red}{31.01} / \textcolor{blue}{0.8775} / \textcolor{blue}{0.47}      \\
     \bottomrule
     
     \midrule
   \multirow{3}{0.8cm}{B100} & $\times2$  & 31.24 / 0.8881 / 3.81 & \textcolor{red}{31.89} / \textcolor{blue}{0.8956} / \textcolor{red}{0.17} & 30.61 / 0.8880 / 1.09  & \textcolor{blue}{31.86} /  \textcolor{red}{0.8959} / 0.52 \\
     & $\times3$  & 28.40 / 0.7869 / 3.74 & \textcolor{red}{28.76} / \textcolor{red}{0.7973} / \textcolor{red}{0.17} & 27.47 / 0.7851 / 1.14  & \textcolor{red}{28.76} / \textcolor{blue}{0.7966} / \textcolor{blue}{0.49}    \\
     & $\times4$  & 26.88 / 0.7114 / 3.32 & \textcolor{blue}{27.18} / \textcolor{red}{0.7245} / \textcolor{red}{0.17} & 26.06 / 0.7117 / 1.07  & \textcolor{red}{27.21} / \textcolor{blue}{0.7242} / \textcolor{blue}{0.53}     \\
     \bottomrule
  \end{tabular} 
\end{table*}

\begin{figure*}
 \caption{\small \textbf{Super-resolution:} Visual results of one image from Set 5 with a scale factor of $\times3$ along with PSNR(dB) / SSIM. ROIs show the improved human visual results using RC-Net with BN by restoring sharp and clean contour of the hair. When comparing PSNR and SSIM, RC-Net without BN \textit{outperforms} VDSR. }
  \includegraphics[width=\textwidth]{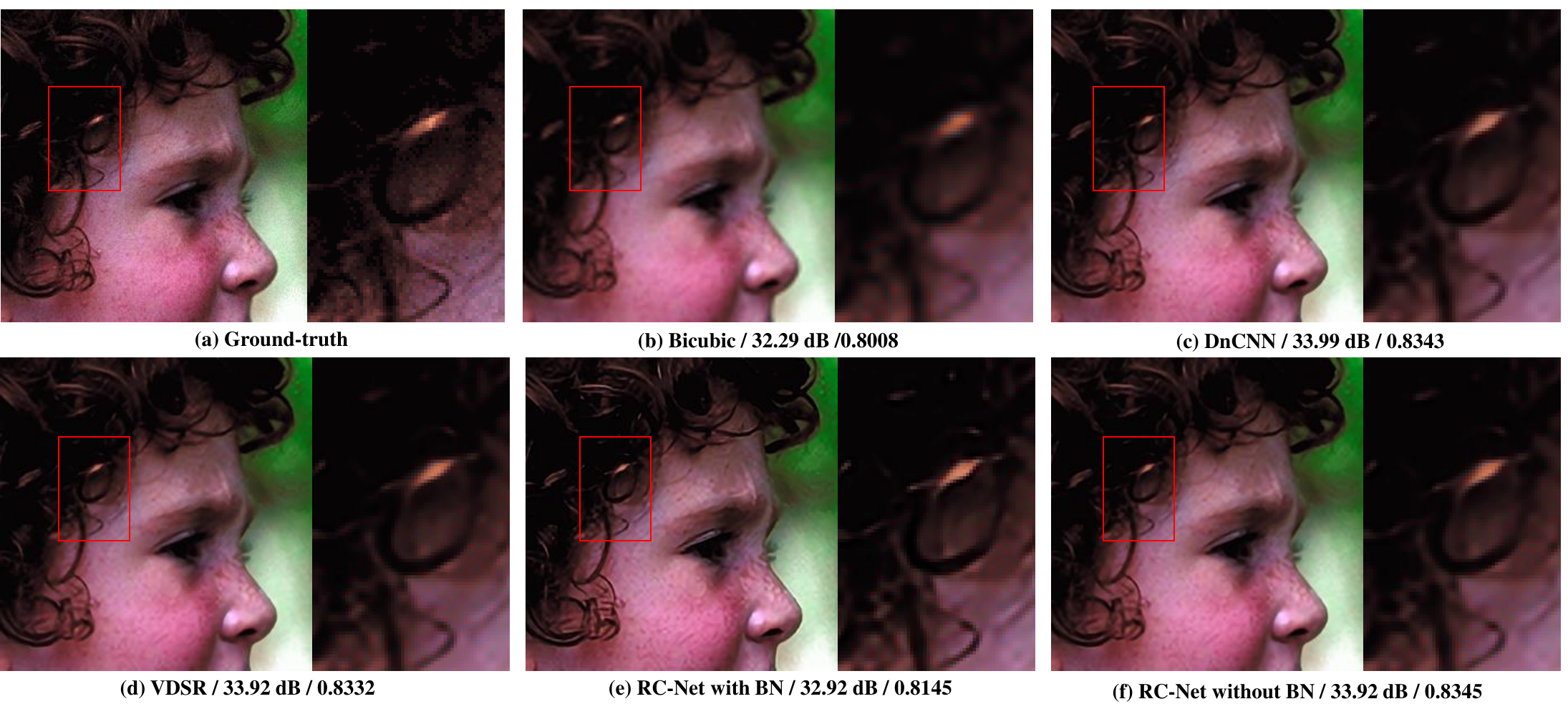}
\label{fig:noBN1}
\end{figure*}

\begin{figure*}
  \caption{\small \textbf{Super-resolution:} Visual results of one image from B100 (100 images from BSD200-test dataset) with scale factor of $\times3$ along with PSNR(dB) / SSIM. Similarly, RC-Net with BN outperforms other methods in restoring sharp details. RC-Net without BN performs excellently on generating image with improved PSNR and SSIM.}
  \includegraphics[width=\textwidth]{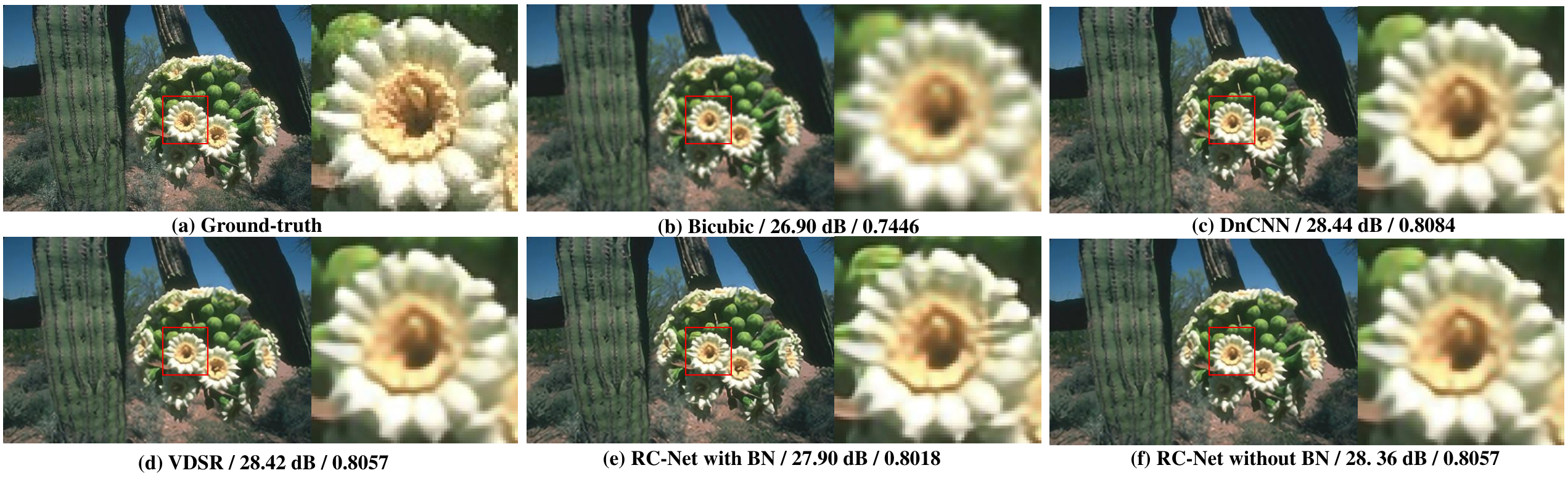}
\label{fig:noBN2}
\end{figure*}
 
 Figure~\ref{fig:noBN1} and Figure\ref{fig:noBN2} are visual results of single image SR with scale factor $\times3$. We removed all BN layers of a RC-Net and call this network as ``RC-Net without BN'', while the original RC-Net keeping all BN layers is called ``RC-Net with BN''. Figure~\ref{fig:noBN1} shows an image from dataset Set5.  The appealing human visual results showing sharp and bright hair recovered by RC-Net with BN. However, removing BN layers leads a better quantitative evaluation in terms of PSNR and SSIM, and comparable performance to the state-of-the-art methods VDSR and DnCNN. We illustrate one more SR result from the dataset BSD200-test shown in Figure \ref{fig:noBN2}. Simiarly, RC-Net with BN restores sharp edges, while RC-Net without BN network performs better in terms of PSNR and SSIM.
\section{More Discussions}
\paragraph{The effect of Batch Normalization layer.}
Normally, BN layers are introduced into the network for reducing internal covariate shift and then accelerating training speed. However, BN layers are preseted as a regulator in our RC-Nets by potentially storing and enhancing high-frequency features. This property helps with restoring images to appear visually close to ground-truth. RC-Nets with BN layer performs remarkably well in the denoising task. In the SR task, RC-Nets with BN provides nice human visual results while RC-Nets without BN achieve excellent quantitative numbers. 

\paragraph{The properties of large-small filter combination.} Indeed, the large-small filter combination introduced in our RC-Nets fundamentally strengthens our model in image restoration. High-frequency features collected by large filters provide major features while details collected by small filters help to regulate the learned feature maps. After removing BN layers, the competitive PSNR and SSIM values demonstrate that our large-small filter combination can enhance RC-Nets' generalization ability both visually and quantitatively.

\end{document}